\def\year{2017}\relax
\documentclass[letterpaper]{article}
\usepackage{aaai17}
\usepackage{times}
\usepackage{helvet}
\usepackage{courier}
\usepackage{latexsym}
\usepackage{amsmath}
\usepackage{multirow}
\usepackage{graphicx}
\begin{document}
%
\title{Definition Modeling: Learning to define word embeddings in natural language}
\author{Thanapon Noraset, Chen Liang, Larry Birnbaum, \and Doug Downey\\
Department of Electrical Engineering \& Computer Science \\
Northwestern University, Evanston IL 60208, USA\\
\{nor, chenliang2013\}@u.northwestern.edu, \{l-birnbaum,d-downey\}@northwestern.edu
}
\maketitle
\begin{abstract}
Distributed representations of words have been shown to capture lexical semantics, as demonstrated by their effectiveness in word similarity and analogical relation tasks.  But, these tasks only evaluate lexical semantics indirectly. In this paper, we study whether it is possible to utilize distributed representations to generate dictionary definitions of words, as a more direct and transparent representation of the embeddings' semantics. We introduce {\em definition modeling}, the task of generating a definition for a given word and its embedding.  We present several definition model architectures based on recurrent neural networks, and experiment with the models over multiple data sets. Our results show that a model that controls dependencies between the word being defined and the definition words performs significantly better, and that a character-level convolution layer designed to leverage morphology can complement word-level embeddings. Finally, an error analysis suggests that the errors made by a definition model may provide insight into the shortcomings of word embeddings.
\end{abstract}


\section{Introduction}
\label{sec-intro}

Distributed representations of words, or word {\em embeddings}, are a key component in many natural language processing (NLP) models \cite{turian_word_2010,huang2014learning}. Recently, several neural network techniques have been introduced to learn high-quality word embeddings from unlabeled textual data \cite{mikolov_efficient_2013,pennington_glove_2014,yogatama_learning_2015}. Embeddings have been shown to capture lexical syntax and semantics.  For example, it is well-known that nearby embeddings are more likely to represent synonymous words \cite{landauer1997solution} or words in the same class \cite{downey2007sparse}.  More recently, the vector offsets between embeddings have been shown to reflect analogical relations \cite{mikolov_linguistic_2013}. However, tasks such as word similarity and analogy only evaluate an embedding's lexical information indirectly.

In this work, we study whether word embeddings can be used to generate natural language definitions of their corresponding words. Dictionary definitions serve as direct and explicit statements of word meaning.  Thus, compared to the word similarity and analogical relation tasks, definition generation can be considered a more transparent view of the syntax and semantics captured by an embedding. We introduce \textit{definition modeling}: the task of estimating the probability of a textual definition, given a word being defined and its embedding. Specifically, for a given set of word embeddings, a definition model is trained on a corpus of word and definition pairs. The models are then tested on how well they model definitions for words not seen during the training, based on each word's embedding. 

\begin{table}
\centering
  \begin{tabular}{ l  p{5.3cm} }
  	\hline
    Word & Generated definition   \\ \hline
    brawler & a person who fights \\ 
    butterfish & a marine fish of the atlantic coast \\
    continually & in a constant manner \\
    creek & a narrow stream of water \\
    feminine & having the character of a woman \\ 
    juvenility & the quality of being childish \\ 
    mathematical & of or pertaining to the science of \\ 
    & mathematics \\
    negotiate & to make a contract or agreement \\ 
    prance & to walk in a lofty manner \\ 
    resent & to have a feeling of anger or dislike \\ 
    similar & having the same qualities \\ 
    valueless & not useful \\ 
     \hline
  \end{tabular}
\caption{Selected examples of generated definitions. The model has been trained on occurrences of each example word in running text, but not on the definitions.}
\label{tab-ex-def}
\end{table}

The definition models studied in this paper are based on recurrent neural network (RNN) models \cite{elman_finding_1990,hochreiter_long_1997}. RNN models have established a new state-of-the-art performance on many sequence prediction and natural language generation tasks \cite{cho_learning_2014,sutskever_sequence_2014,karpathy_deep_2014,wen_semantically_2015}. An important characteristic of dictionary definitions is that only a subset of the words in the definition depend strongly on the word being defined.  For example, the word ``woman'' in the definition of ``feminine" in Table \ref{tab-ex-def} depends on the word being defined than the rest. To capture the varying degree of dependency, we introduce a gated update function that is trained to control information of the word being defined used for generating each definition word. Furthermore, since the morphemes of the word being defined plays a vital role in the definition, we experiment with a character-level convolutional neural network (CNN) to test whether it can provide complementary information to the word embeddings. Our best model can generate fluent and accurate definitions as shown in Table \ref{tab-ex-def}. We note that none of the definitions in the table exactly match any definition seen during training.

Our contributions are as follows: (1) We introduce the definition modeling task, and present a probabilistic model for the task based on RNN language models. (2) In experiments with different model architectures and word features, we show that the gate function improves the perplexity of a RNN language model on definition modeling task by ${\sim}10\%$, and the character-level CNN further improves the perplexity by ${\sim}5\%$. (3) We also show that the definition models can be use to perform the reverse dictionary task studied in previous work, in which the goal is to match a given definition to its corresponding word. Our model achieves an 11.8\% absolute gain in accuracy compared to previous state-of-the-art by Hill et al. \shortcite{hill_learning_2016}. (4) Finally, our error analysis shows that a well-trained set of word embeddings pays significant role in the quality of the generated definitions, and some of error types suggest shortcomings of the information encoded in the word embeddings.


\section{Previous Work}

Our goal is to investigate RNN models that learns to define word embeddings by training on examples of dictionary definitions. While dictionary corpora have been utilized extensively in NLP, to the best of our knowledge none of the previous work has attempted create a generative model of definitions. Early work focused on {\em extracting} semantic information from definitions. For example, Chodorow \shortcite{chodorow1985extracting}, and Klavans and Whitman \shortcite{klavans2001extracting} constructed a taxonomy of words from dictionaries. Dolan et al. \shortcite{dolan1993automatically} and Vanderwende et al. \shortcite{vanderwende2005} extracting semantic representations from definitions, to populate a lexical knowledge base. 

In distributed semantics, words are represented by a dense vector of real numbers, rather than semantic predicates.  Recently, dictionary definitions have been used to learn such embeddings. For example, Wang et al. \shortcite{wang_learning_2015} used words in definition text as a form of ``context'' words for the Word2Vec algorithm \cite{mikolov_distributed_2013}. Hill et al. \shortcite{hill_learning_2016} use dictionary definitions to model compositionality, and evaluate the models with the reverse dictionary task.  While these works learn word or phrase embeddings from definitions, we only focus on generating definitions from existing (fixed) embeddings. Our experiments show that our models outperform those of Hill et al. \shortcite{hill_learning_2016} on the reverse dictionary task. 

Our work employs embedding models for natural language generation.  A similar approach has been taken in a variety of recent work on tasks distinct from ours. Dinu and Baroni \shortcite{dinu_how_2014} present a method that uses embeddings to map individual words to longer phrases denoting the same meaning.  Likewise, Li et al. \shortcite{Li2015AHN} study how to encode a paragraph or document as an embedding, and reconstruct the original text from the encoding.  Other recent work such as the image caption generation \cite{karpathy_deep_2014} and spoken dialog generation \cite{wen_semantically_2015} are also related to our work, in that a sequence of words is generated from a single input vector. Our model architectures are inspired by sequence-to-sequence models \cite{cho_learning_2014,sutskever_sequence_2014}, but definition modeling is distinct, as it is a {\em word}-to-sequence task.


\section{Dictionary Definitions}
\label{sec-definition}

In this section, we first investigate definition content and structure through a study of existing dictionaries.  We then describe our new data set, and define our tasks and metrics.

\subsection{Definition Content and Structure}
In existing dictionaries, individual definitions are often comprised of \textit{genus} and \textit{differentiae} \cite{chodorow1985extracting,montemagni1992structural}. The {\em genus} is a generalized class of the word being defined, and the {\em differentiae} is what makes the word distinct from others in the same class. For instance, 
\begin{quote}
\textbf{Phosphorescent}: emitting light without appreciable heat 
as by slow oxidation of phosphorous
\end{quote}
``emitting light'' is a genus, and ``without applicable heat ...'' is a differntiae. Furthermore, definitions tend to include common patterns such as ``the act of ...'' or ``one who has ...'' \cite{markowitz1986semantically}. However, the patterns and styles are often unique to each dictionary.

The \textit{genus + differentiae (G+D)} structure is not the only pattern for definitions. For example, the entry below exhibits distinct structures.
\begin{quote}
\textbf{Eradication}: the act of plucking up by the roots; a rooting out; extirpation; utter destruction
\end{quote}
This set of definitions includes a synonym (``extirpation''), a reverse of the \textit{G+D} structure (``utter destruction''), and an uncategorized structure (``a rooting out''). 

\subsection{Corpus: Preprocessing and Analysis}
\label{sec-definition-analysis}
Dictionary corpora are available in a digital format, but are designed for human consumption and require preprocessing before they can be utilized for machine learning. Dictionaries contain non-definitional text to aid human readers, e.g. the entry for ``gradient'' in Wordnik\footnote{https://www.wordnik.com/words/gradient} contains fields (``Mathematics'') and example usage (``as, the gradient line of a railroad.''). Further, many entries contain multiple definitions, usually (but not always) separated by ``;''.

We desire a corpus in which each entry contains only a word being defined and a single definition. We parse dictionary entries from GCIDE\footnote{http://gcide.gnu.org.ua/} and preprocess WordNet's glosses, and the fields and usage are removed. The parsers and preprocessing scripts can be found at https://github.com/northanapon/dict-definition.

To create a corpus of reasonable size for machine learning experiments, we sample around 20k words from the 50k most frequent words in the Google Web 1T corpus \cite{brants2006web}, removing function words. In addition, we limit the number of entries for each word in a dictionary to three before the splitting by ``;'' (so that each word being defined may repeat multiple times in our corpus). After cleaning and pruning, the corpus has a vocabulary size of 29k.  Other corpus statistics are shown in Table \ref{tab-stat}.
\begin{table}
\centering
  \begin{tabular}{ r || r | r | r }
  	\hline
    Split & train & valid & test \\ \hline
    \#Words &  20,298 & 1,127 & 1,129\\ 
    \#Entries & 146,486 & 8,087 & 8,352 \\ 
    \#Tokens & 1,406,440 & 77,948 & 79,699\\
    Avg length & 6.60 & 6.64 & 6.54 \\
    \hline
  \end{tabular}
\caption{Basic statistics of the common word definitions corpus. 
Splits are mutually exclusive in the words being defined.}
\label{tab-stat}
\end{table}

We analyze the underlying structure of the definitions in the corpus by manually 
labeling each definition with one of four structures:
\textit{G+D}, \textit{D+G}, \textit{Syn} (synonym), and \textit{Misc}. 
In total, we examine 680 definitions from 100 randomly selected words. 
The results are shown in Table \ref{tab-struct_analysis}.
We reaffirm earlier studies showing that the \textit{G+D} structure dominates 
in both dictionaries.
However, other structures are also present, highlighting the challenge inherent in the dictionary modeling task. Further, we observe that the \textit{genus} term is sometimes general (e.g., ``one'' or ``that''), and other times specific (e.g. ``an advocate''). 

\begin{table}
 \centering
  \begin{tabular}{ r || r | r | l }
  	\hline
    Label & WN & GC & Example \\ \hline
    \textit{G+D} & 85\% & 50\% & to divide into thin plates\\ 
    \textit{D+G} & 7\% & 9\% & a young deer  \\ 
    \textit{Syn} & 1\% & 32\% & energy \\ 
    \textit{Misc.} & 4\% & 8\% & in a diagonal direction  \\ 
    \textit{Error} & 3\% & 1\% & used as intensifiers \\  \hline 
    \textbf{Total} & 256 & 424 & \\ \hline
  \end{tabular}
\caption{The number of manually labeled structures of dictionary definitions in WordNet (WN) 
and GCIDE (GC). \textit{G+D} is \textit{genus} followed by \textit{differentiae}, and 
\textit{D+G} is the reverse. \textit{Syn} is a synonym. The words defined in the example column are
``laminate'', ``fawn'', ``activity'', ``diagonally'', and ``absolutely''.}
\label{tab-struct_analysis}
\end{table}

\subsection{Dictionary Definition Tasks}
\label{sec-dict-task}

In the definition modeling (DM) task, we are given an input word $w^*$, and output the likelihood of any given text $D$ being a definition of the input word.  In other words, we estimate $P(D|w^*)$.  We assume our definition model has access to a set of word embeddings, estimated from some corpus other than the definition corpus used to train the definition model.

DM is a special case of language modeling, and as in language modeling the performance of a definition model can be measured by using the perplexity of a test corpus. Lower perplexity suggests that the model is more accurate at capturing the definition structures and the semantics of the word being defined. 

Besides perplexity measurement, there are other tasks that we can use to further evaluate a dictionary definition model including definition generation, and the reverse and forward dictionary tasks. In definition generation, the model produces a definition of a test word.  In our experiments, we evaluate generated definitions using both manual examination and BLEU score, an automated metric for generated text. The reverse and forward dictionary tasks are ranking tasks, in which the definition model ranks a set of test words based on how likely they are to correspond to a given definition (the {\em reverse dictionary} task) or ranks a set of test definitions for a given word (the {\em forward dictionary} task) \cite{hill_learning_2016}. A dictionary definition model achieves this by using the predicted likelihood $P(D|w^*)$ as a ranking score.


\section{Models}
\label{sec-model}

The goal of a definition model is to predict the probability of a definition ($D = [w_1,...,w_T]$) given a word being defined $w^*$. Our model assumes that the probability of generating the $t$th word $w_t$ of a definition text depends on both the previous words and the word being defined (Eq \ref{eq-main}).  The probability distribution is usually  approximated by a softmax function (Eq \ref{eq-softmax})
\begin{align}
p(D|w^*) = \prod_{t=1}^{T} p(w_t|w_1,..,w_{t-1},w^*) \label{eq-main}\\
p(w_t=j|w_1,..,w_{t-1},w^*) \propto e^{W_jh_t/\tau} \label{eq-softmax}
\end{align}
where $W_j$ is a matrix of parameters associated with word $j$, $h_t$ is a vector summarizing inputs so far at token $t$, and $\tau$ is a hyper-parameter for temperature, set to be 1 unless specified. Note that in our expression, the word being defined $w^*$ is present at all time steps as an additional conditioning variable.

The definition models explored in this paper are based on a recurrent neural network language model (RNNLM) \cite{mikolov_recurrent_2010}. An RNNLM is comprised of RNN units, where each unit reads one word $w_t$ at every time step $t$ and outputs a hidden representation $h_t$ for Eq \ref{eq-softmax}.
\begin{equation}
h_t = g(v_{t-1}, h_{t-1}, v^*)
\label{eq-hidden}
\end{equation}
where $g$ is a recurrent nonlinear function, $v_t$ denotes the embedding (vector representation) of the word $w_t$, and $v^*$ is likewise the embedding of the word being defined.

\subsection{Model Architectures}
\label{sec-model-ri}

A natural method to condition an RNN language model is to provide the network with the word being defined at the first step, as a form of ``seed'' information.  The seed approach has been shown to be effective in RNNs for other tasks \cite{kalchbrenner_recurrent_2013,karpathy_deep_2014}. Here, we follow the simple method of Sutskever et al., \shortcite{sutskever_generating_2011}, in which the seed is added at the beginning of the text. In our case, the word being defined is added to the beginning of the definition. Note that we ignore the predicted probability distribution of the seed itself at test time.

Section \ref{sec-definition} shows that definitions exhibit common patterns. We hypothesize that the word being defined should be given relatively more important for portions of the definition that carry semantic information, and less so for patterns or structures comprised of function and stop words.  Further, Wen et al. \shortcite{wen_stochastic_2015} show that providing constant seed input at each time step can worsen the overall performance of spoken dialog generators.

Thus, inspired by the GRU update gate \cite{cho_learning_2014}, we update the output of the recurrent unit with GRU-like update function as:
\begin{align}
z_t &= \sigma(W_z[v^*;h_t]+b_z) \label{eq-z}\\
r_t &= \sigma(W_r[v^*;h_t]+b_r) \label{eq-r}\\
\tilde{h_t} &= tanh(W_h[(r_t\odot{}v^*);h_t] + b_h) \label{eq-grulike-newh}\\
h_t &= (1-z_t)\odot{}h_t + z_t \odot{} \tilde{h_t} \label{eq-grulike}
\end{align}
where $\sigma$ is the sigmoid function, $[a;b]$ denotes vector concatenation, and $\odot{}$ denotes element-wise multiplication. $h_t$ from Eq \ref{eq-hidden} is updated as given in Eq \ref{eq-grulike}. At each time step, $z_t$ is an \textit{update} gate controlling how much the output from RNN unit changes, and $r_t$ is a \textit{reset} gate controlling how much information from the word being defined is allowed. We name this model \textit{Gated} (\textit{G}).

In the rest of this subsection, we present three baseline model architectures that remove portions of \textit{Gated}.  In our experiments, we will compare  the performance of \textit{Gated} against the baselines in order to measure the contribution of each portion of our architecture. First, we reduce the model into a standard RNNLM, where 
\begin{equation}
h_t = g(v_{t-1}, h_{t-1})
\label{eq-standardlm}
\end{equation}
The standard model {\em only} receives information about $w^*$ at the first step (from the seed).  We refer to this baseline as \textit{Seed} (\textit{S}).

A straightforward way to incorporate the word being defined throughout the definition is simply to provide its embedding $v^*$ as a constant input at every time step.  We refer to this model as \textit{Input} (\textit{I}):
\begin{equation}
h_t = g([v^*;v_{t-1}], h_{t-1}) \label{eq-inputrnn}
\end{equation}
\cite{mikolov_context_2012}.
Alternatively, the model could utilize $v^*$ to update the hidden representation from the RNN unit, named \textit{Hidden} (\textit{H}). The update function for \textit{Hidden} is:
\begin{equation}
h_t = tanh(W_h[v^*;h_t] + b_h)
\label{eq-concat-update}
\end{equation}
where $W_h$ is a weight matrix, and $b_h$ is the bias. In \textit{Hidden} we update $h_t$ from Eq \ref{eq-hidden} using Eq \ref{eq-concat-update}.  This is similar to the GRU-like architecture in Eq \ref{eq-grulike} without the gates (i.e. $r_t$ and $z_t$ are always vectors of 1s).

\subsection{Other Features}
\label{sec-model-features}

In addition to model architectures, we explore whether other word features derived from the word being defined can provide complementary information to the word embeddings. We focus on two different features: affixes, and hypernym embeddings. To add these features within DM, we simply concatenate the embedding $v^*$ with the additional feature vectors. 

\subsubsection*{Affixes}
Many words in English and other languages consist of composed morphemes. For example, a word ``capitalist'' contains a root word ``capital'' and a suffix ``-ist''. A model that knows the semantics of a given root word, along with knowledge of how affixes modify meaning, could accurately define any morphological variants of the root word. However, automatically decomposing words into morphemes and deducing the semantics of affixes is not trivial.

We attempt to capture prefixes and suffixes in a word by using character-level information. We employ a character-level convolution network to detect affixes \cite{LeCun_NIPS1989_293}.  Specifically, $w^*$ is represented as a sequence of characters with one-hot encoding. A padding character is added to the left and the right to indicate the start and end of the word. We then apply multiple kernels of varied lengths on the character sequence, and use max pooling to create the final features \cite{kim_AAAI1612489}. We hypothesize that the convolution input, denoted as \textit{CH}, will allow the model to identify regularities in how affixes alter the meanings of words.

\subsubsection*{Hypernym Embeddings}
As we discuss in Section \ref{sec-definition}, dictionary definitions often follow a structure of \textit{genus} + \textit{differentia}. We attempt to exploit this structure by providing the model with knowledge of the proper \textit{genus}, drawn from a database of hypernym relations. In particular, we obtain the hypernyms from WebIsA database \cite{seitnerlarge} which employs Hearst-like patterns \cite{hearst_automatic_1992} to extract hypernym relations from the Web.  We then provide an additional input vector, referred to as \textit{HE}, to the model that is equal to the weighted sum of the top $k$ hypernyms in the database for the word being defined.  In our experiments $k=5$ and the weight is linearly proportional to the frequency in the resource.  For example, the top 5 hypernyms and frequencies for ``fawn'' are ``color'':149, ``deer'':135, ``animal'': 132.0, ``wildlife'':82.0, ``young'': 68.0.

\section{Experiments and Results}
\label{sec-exp}

We now present our experiments evaluating our definition models.  We train multiple model architectures using the \textit{train} set and evaluate the model using the \textit{test} set on all of the three tasks described in Section \ref{sec-dict-task}. We use the \textit{valid} set to search for the learning hyper-parameters. Note that the words being defined are mutually exclusive across the three sets, and thus our experiments evaluate how well the models generalize to new words, rather than to additional definitions or senses of the same words.

All of the models utilize the same set of fixed, pre-trained word embeddings from the Word2Vec project,\footnote{https://code.google.com/archive/p/word2vec/} and a 2-layer LSTM network as an RNN component \cite{hochreiter_long_1997}. The embedding and LSTM hidden layers have 300 units each. For the affix detector, the character-level CNN has kernels of length 2-6 and size \{10, 30, 40, 40, 40\} with a stride of 1. During training, we maximize the log-likelihood objective using Adam, a variation of stochastic gradient decent \cite{kingma_adam:_2014}. The learning rate is 0.001, and the training stops after 4 consecutive epochs of no significant improvement in the validation performance. The source code and dataset for our experiment can be found at https://github.com/websail-nu/torch-defseq.

\subsection{Definition Modeling}
\label{sec-exp-lm}

First, we compare our different methods for utilizing the word being defined within the models.  The results are shown in the first section of Table \ref{tab-result_ppl}. We see that the gated update (\textit{S+G}) improves the performance of the \textit{Seed}, while the other architectures (\textit{S+I} and \textit{S+H}) do not. The results are consistent with our hypothesis that the word being defined is more relevant to some words in the definition than to others, and the gate update can identify this.  We explore the behavior of the gate further in Section \ref{sec-disc}.

Next, we evaluate the contribution of the linguistic features. We see that the \textit{affixes} (\textit{S+G+CH}) further improves the model, suggesting that character-level information can complement word embeddings learned from context. Perhaps surprisingly, the \textit{hypernym embeddings} (\textit{S+G+CH+HE}) have an unclear contribution to the performance. We suspect that the average of multiple embeddings of the hypernym words may be a poor representation the \textit{genus} in a definition. More sophisticated methods for harnessing hypernyms are an item of future work.

\begin{table}
\centering
  \begin{tabular}{ l | r | r }
  	\hline
    Model & \#Params & Perplexity  \\ \hline \hline
    \textit{Seed} & 10.2m & 56.350\\ 
    \textit{S+I} & 10.6m & 57.372  \\ 
    \textit{S+H} & 10.4m & 58.147  \\ 
    \textit{S+G} & 10.8m & 50.949 \\ \hline
    \textit{S+G+CH} & 11.1m & 48.566 \\ 
    \textit{S+G+CH+HE} & 11.7m & \textbf{48.168} \\
    \hline
    \hline
  \end{tabular}
\caption{Perplexity evaluated on dictionary entries in the test set (lower is better). 
}
\label{tab-result_ppl}
\end{table}

\subsection{Definition Generation}
In this experiment, we evaluate the quality of the definitions generated by our models.  We compute BLEU score between the output definitions and the dictionary definitions to  measure the quality of the generation. The decoding algorithm is simply sampling a token at a time from the model's predicted probability distribution of words. We sample 40 definitions for each word being defined, using a temperature ($\tau$ in Eq \ref{eq-softmax}) that is close to a greedy algorithm (0.05 or 0.1, selected from the \textit{valid} set) and report the average BLEU score. For help in interpreting the BLEU scores, we also report the scores for three baseline methods that output definitions found in the training or test set.  The first baseline, \textit{Inter}, returns the definition of the test set word from the other dictionary.  Its score thus reflects that of a definition that is semantically correct, but differs stylistically from the target dictionary.  The other baselines (\textit{NE-WN} and \textit{NE-GC}) return the definition from the training set for the embedding nearest to that of the word being defined. In case of a word having multiple definitions, we micro-average BLEU scores before averaging an overall score.

\begin{table}
\centering
  \begin{tabular}{ l || r | r | r | r }
  	\hline
	Model & GC & WN & Avg & Merged   \\ \hline \hline
	\textit{Inter}  & 27.90 & 21.15 & - & - \\ 
	\textit{NE} & 29.56 & 21.42 & 25.49 & 34.51\\
    \textit{NE-WN} & 22.70 & \textbf{27.42} & 25.06 & 32.16 \\ 
    \textit{NE-GC} & \textbf{33.22} & 17.77 & 25.49 & 35.45 \\
	 \hline
	\textit{Seed} & 26.69 & 22.46 & 24.58 & 30.46 \\
    \textit{S+I} & 28.44 & 21.77 & 25.10 & 31.58 \\
    \textit{S+H} & 27.43 & 18.82 & 23.13 & 29.66 \\
    \textit{S+G} & 30.86 & 23.15 & 27.01 & 34.72 \\ \hline
    \textit{S+G+CH} & 31.12 & 24.11 & \textbf{27.62} & \textbf{35.78} \\
    \textit{S+G+CH+HE} & 31.10 & 23.81 & 27.46 & 35.28 \\ \hline 
    \multicolumn{4}{l}{Additional experiments} \\ \hline
    \textit{Seed*} & 27.24 & 22.78 & 25.01 & 31.15\\
    \textit{S+G+CH+HE*} & 33.39 & 25.91 & 29.65 & 38.35 \\
    Random Emb &  22.09 & 20.05 & 21.07 & 24.77 \\
    \hline
    \hline
  \end{tabular}
\caption{Equally-weighted BLEU scores for up to 4-grams, on definitions evaluated using different reference dictionaries (results are not comparable between columns).} 
\label{tab-result_bleu}
\end{table}

Table \ref{tab-result_bleu} shows the BLEU scores of the generated definitions given different reference dictionaries. AVG and Merge in the table are two ways of aggregating the BLEU score. AVG averages the BLEU scores by using each dictionary as the ground truth. The Merge computes score by using union of the two dictionaries. First, we can see that the baselines have low BLEU scores when evaluated on definitions from the other dictionary (\textit{Inter} and \textit{NE-}). This shows that different dictionaries use different styles. However, despite the fact that our best model \textit{S+G+CH} is unaware of which dictionary it is evaluated against, it generates definitions that strike a balance between both dictionaries, and achieves higher BLEU scores overall. As in the earlier experiments, the \textit{Gated} model improves the most over the \textit{Seed} model. In addition, the \textit{affixes} further improves the performance while the contribution of the \textit{hypernym embeddings} is unclear on this task. 

It is worth noting that many generated definitions contain a repeating pattern (i.e. ``a metal, or other materials, or other materials''). We take the definitions from the language model (\textit{Seed}) and our full system (\textit{S+G+CH+HE}), and clean the definitions by retaining only one copy of the repeated phrases. We also only output the most likely definition for each word. The BLEU score increases by 2 (\textit{Seed*} and \textit{S+G+CH+HE*}). We discuss about further analysis and common error types in Section \ref{sec-disc}.

\begin{figure*}
\centering
\begin{minipage}{0.45\textwidth}
\centering
\includegraphics[width=0.9\linewidth]{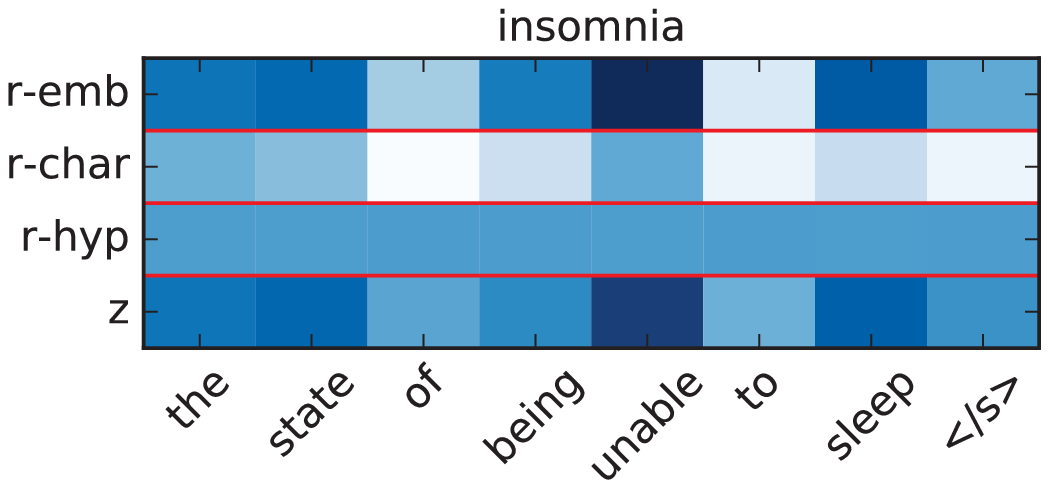}
\end{minipage}
\begin{minipage}{0.45\textwidth}
\centering
\includegraphics[width=0.9\linewidth]{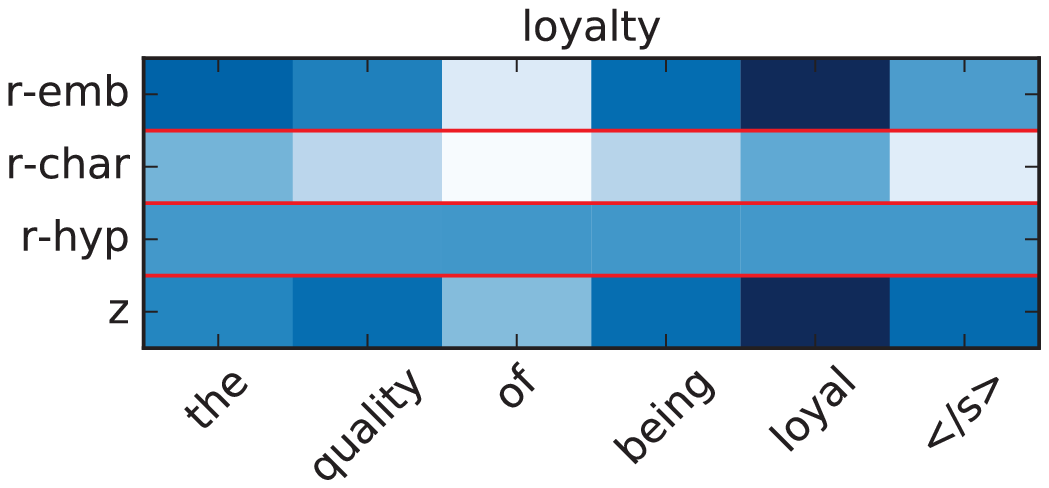}
\end{minipage}
\caption{Average gate activations for tokens of two definitions (omitting seed). The model utilizes the word being defined more for predicting content words than for function words. 
}
\label{fig-activation}
\end{figure*}

\subsection{Reverse and Forward Dictionary}

\begin{table}
\centering
  \begin{tabular}{l | r || r | r || r }
  	\hline
    \multirow{2}{*}{Model} & \multirow{2}{*}{\#Params} & \multicolumn{2}{c||}{RVD}  & \multicolumn{1}{c}{FWD}   \\
    \cline{3-5}
     & & @1 & @10 & R-Prec  \\ \hline \hline
        \textit{BOW w2v cosine} & 0.09m & 0.106 & 0.316  & -\\
    \textit{RNN w2v cosine} & 1.82m & 0.190 & 0.452 & - \\ \hline
    \textit{Seed} & 10.2m & 0.175 & 0.465 & 0.163 \\ 
    \textit{S+I} & 10.6m & 0.187 & 0.492 & 0.169 \\ 
    \textit{S+H} & 10.4m & 0.286 & 0.573 & 0.282  \\ 
    \textit{S+G} & 10.8m & 0.293 & 0.581 & 0.282\\ \hline
    \textit{S+G+CH} & 11.1m & 0.307 & 0.600 & 0.298 \\ 
    \textit{S+G+CH+HE} & 11.7m & \textbf{0.308} & \textbf{0.608} & \textbf{0.304} \\
    \hline
    \hline
  \end{tabular}
\caption{Model performance on Reverse (RVD) and Forward (FWD) Dictionary tasks 
}
\label{tab-result_dict}
\end{table}

In the dictionary tasks, the models are evaluated by how well they rank words for given definitions (RVD) or definitions for words (FWD).  We compare against models from previous work on the reverse dictionary task \cite{hill_learning_2016}. The previous models read a definition and output an embedding, then use cosine similarity between the output embedding and the word embedding as a ranking score. There are two ways to compose the output embedding: \textit{BOW w2v cosine} uses vector addition and linear projection, and \textit{RNN w2v cosine} uses a single-layer LSTM with 512 hidden units. We use two standard metrics for ranked results, accuracy at top $k$ and $R$-Precision (i.e. precision of the top $R$ where $R$ is the number of correct definitions for the test word).

Table \ref{tab-result_dict} shows that our models perform well on the dictionary tasks, even though they are trained to optimize a distinct objective (definition likelihood). However, we note that our models have more parameters than those from previous work. Furthermore, we find that \textit{RNN w2v cosine} performs better than \textit{BOW w2v cosine}, which differs from the previous work. The differences may arise from our distinct preprocessing described in Section \ref{sec-definition}, i.e. redundant definitions are split into multiple definitions. We omit the information retrieval approach baseline because it is not obvious how to search for unseen words in the test set.


\section{Discussion}
\label{sec-disc}

In this section, we discuss our analysis of the generated definitions. We first present a qualitative evaluation, followed by an analysis on how the models behave. Finally, we discuss error types of the generated definitions and how it might reflect information captured in the word embeddings.

\subsection{Qualitative Evaluation and Analysis}

First, we perform a qualitative evaluation of the models' output by asking 6 participants to rank a set of definitions of 50 words sampled from the test set. For each word $w$, we provide in random order: a ground-truth definition for $w$ (\textit{Dictionary}), a ground-truth definition of the word $w'$ whose embedding is nearest to that of $w$ (\textit{NE}), the standard language model (\textit{Seed*}), and our full system (\textit{S+G+CH+HE*}). Inter-annotator agreement was strong (almost all inter-annotator correlations were above 0.6).  Table \ref{tab-survey} shows that definitions from the \textit{S+G+CH+HE*} are ranked second after the dictionary definitions, on average. The advantage of \textit{S+G+CH+HE*} over \textit{Seed*} is statistically significant ($p < 0.002$, t-test), and the difference between \textit{S+G+CH+HE*} is and \textit{NE} is borderline significant ($p < 0.06$, t-test).
\begin{table}
\centering
  \begin{tabular}{ l || r | r | r | r | r }
  	\hline
	Choices & @1 & @2 & @3 & @4 & Avg   \\ \hline \hline
	Dictionary & 58.3 & 21.9 & 7.72 & 10.1 & 1.64 \\
    \textit{NE} & 16.3 & 22.8 & 27.85 & 37.0 & 2.75 \\
	\textit{Seed*} & 6.8 & 23.5 & 35.23 & 35.1 & 2.92 \\
    \textit{S+G+CH+HE*} & 18.7 & 31.8 & 29.19 & 17.8 & 2.41 \\
    \hline
    \hline
  \end{tabular}
\caption{Percentage of times a definition is manually ranked in each position (@k), and average rank (Avg).} 
\label{tab-survey}
\end{table}

All of our results suggest that the gate-based models are more effective.  We investigate this advantage by plotting the average gate activation ($z$ and $r$ in Eq \ref{eq-z} and \ref{eq-r}) in Figure \ref{fig-activation}. The $r$ gate is split into 3 parts corresponding to the embedding, character information, and the hypernym embedding. The figure shows that the gate makes the output distribution more dependent on the word being defined when predicting content words, and less so for function words. The hypernym embedding does not contribute to the performance and its gate activation is relatively constant. Additional examples can be found in the supplementary material.

Finally, we present a comparison of definitions generated from different models to gain a better understanding of the models. Table \ref{tab-def-progress} shows the definitions of three words from Table \ref{tab-ex-def}. The \textit{Random Embedding} method does not generate good definitions.  The nearest embedding method \textit{NE} returns a similar definition, but often makes important errors (e.g., ``feminine'' vs ``masculine''). The models using the gated update function generate better definitions, and the character-level information is often informative for selecting content words (e.g. ``mathematics'' in ``mathematical'').

\begin{table*}
\centering
  \begin{tabular}{ l | p{3.5cm} | p{4.6cm} | p{5.7cm} }
  	\hline
    Model & creek & feminine & mathematical \\ \hline \hline
	\textit{Random Emb} & to make a loud noise 	& to make a mess of & of or pertaining to the middle \\ \hline
    \textit{NE} & any of numerous bright translucent organic pigments & a gender that refers chiefly but not exclusively to males or to objects classified as male & of or pertaining to algebra \\ \hline \hline
    \textit{Seed} & a small stream of water 	& of or pertaining to the fox & of or pertaining to the science of algebra \\ \hline
    \textit{S+I} & a small stream of water 	& of or pertaining to the human body & of or relating to or based in a system \\ \hline
    \textit{S+H} & a stream of water 	& of or relating to or characteristic of the nature of the body & of or relating to or characteristic of the science \\ \hline
    \textit{S+G} & a narrow stream of water 	& having the nature of a woman & of or pertaining to the science \\ \hline \hline
    \textit{S+G+CH} & a narrow stream of water 	& having the qualities of a woman & of or relating to the science of mathematics \\ \hline
    \textit{S+G+CH+HE} & a narrow stream of water 	& having the character of a woman & of or pertaining to the science of mathematics \\ \hline \hline
  \end{tabular}
\caption{Selected examples of generated definitions from different models. We sample 40 definitions for each word and rank them by the predicted likelihood. Only the top-ranked definitions are shown in this table.}
\label{tab-def-progress}
\end{table*}

\subsection{Error Analysis}
In our manual error analysis of 200 labeled definitions. We find that 140 of them contain some degree of error. Table \ref{tab-error} shows the primary error types, with examples. Types (1) to (3) are fluency problems, and likely stem from the definition model, rather than shortcomings in the embeddings.

We believe the other error types stem more from semantic gaps in the embeddings than from limitations in the definition model.  Our reasons for placing the blame on the embeddings rather than the definition model itself are twofold.  First, we perform an ablation study in which we train \textit{S+G+CH} using randomized embeddings, rather than the learned Word2Vec ones. The performance of the model is significantly worsened (the test perplexity is 100.43, and the  BLEU scores are shown in Table \ref{tab-result_bleu}), which shows that the good performance of our definition models is in significant measure due to the embeddings. Secondly, the error types (4) - (6) are plausible shortcomings of embeddings, some of which have been reported in the literature. These erroneous definitions are partially correct (often the correct part of speech), but are missing details that may not appear in contexts of the word due to reporting bias \cite{gordon2013reporting}. For example, the word ``captain'' often appears near the word ``ship'', but the {\em role} (as a leader) is frequently implicit. Likewise, embeddings are well-known to struggle in capturing antonym relations \cite{2016arXiv160808940A}, which helps explain the opposite definitions output by our model.
\begin{table}
\centering
  \begin{tabular}{ l p{6cm} }
  	\hline
    Word & Definition   \\ \hline \hline
    \multicolumn{2}{l}{(1) Redundancy and overusing common phrases: 4.28\%} \\  
    propane &	a volatile flammable gas that is used to burn gas \\
	\hline
    \multicolumn{2}{l}{(2) Self-reference: 7.14\%} \\ 
    precise	& to make a precise effort \\
	\hline
    \multicolumn{2}{l}{(3) Wrong part-of-speech: 4.29\%} \\ 
    accused & to make a false or unethical declaration of \\
    \hline
    \multicolumn{2}{l}{(4) Under-specified: 30.00\%} \\  
    captain & a person who is a member of a ship\\
    \hline
	\multicolumn{2}{l}{(5) Opposite: 8.57\%} \\  
    inward & not directed to the center \\
    \hline
    \multicolumn{2}{l}{(6) Close semantics: 22.86\%} \\  
    adorable & having the qualities of a child\\
    \hline
	\multicolumn{2}{l}{(7) Incorrect: 32.14\%} \\  
    incase & to make a sudden or imperfect sound\\
    \hline
    \hline
  \end{tabular}
\caption{Error types and examples.}
\label{tab-error}
\end{table}


\section{Conclusion}
In this work, we introduce the definition modeling task, and investigate whether word embeddings can be used to generate definitions of the corresponding words.  We evaluate different architectures based on a RNN language model on definition generation and the reverse and forward dictionary tasks. We find the gated update function that controls the influence of the word being defined on the model at each time step improves accuracy, and that a character-level convolutional layer further improves performance. Our error analysis shows a well-trained set of word embeddings is crucial to the models, and that some failure modes of the generated definitions may provide insight into shortcomings of the word embeddings. In future work, we plan to investigate whether definition models can be utilized to improve word embeddings or standard language models.


\section{ Acknowledgments}
This work was supported in part by NSF Grant IIS-1351029 and the Allen Institute for Artificial Intelligence.  We thank Chandra Sekhar Bhagavatula for helpful comments.

\fontsize{9.5pt}{10.5pt} 
\selectfont
\bibliography{aaai}
\bibliographystyle{aaai}
\end{document}